\documentclass[10pt,twocolumn,letterpaper]{article}

\usepackage[final,algorithms]{wacv}
\usepackage{xcolor}
\usepackage{soul}
\sethlcolor{yellow}
\usepackage{amsmath,amssymb,bm}
\usepackage{booktabs}
\usepackage{graphicx}
\usepackage{microtype}
\usepackage{xcolor}
\usepackage{algorithm}
\usepackage{algpseudocode}
\usepackage{tikz}
\usetikzlibrary{arrows.meta,positioning,fit,calc}

\definecolor{wacvblue}{rgb}{0.21,0.49,0.74}
\usepackage[breaklinks,colorlinks,allcolors=wacvblue]{hyperref}

\def\confName{WACV}
\def\confYear{2027}

\title{MoRE-GS: Motion-Regime Experts for Future Gaussian Splatting}

\author{
Shima Salehi \qquad Joshua Peeples\\
Department of Electrical and Computer Engineering\\
Texas A\&M University\\
College Station, Texas, USA\\
{\tt\small shima.salehi@tamu.edu \qquad jpeeples@tamu.edu}
}

\begin{document}

\maketitle

\begin{abstract}
Future dynamic novel-view synthesis requires extrapolating multiple motion
regimes within the same scene. A single dynamics model may not simultaneously
represent stationary regions, coordinated motion, local interactions, and
articulated trajectories. We introduce Motion Regime Expert Routing for
Gaussian Splatting (MoRE-GS), a causal, render-aware framework for forecasting
dynamic 3D Gaussian scenes. MoRE-GS freezes a Gaussian representation
reconstructed exclusively from the observed temporal prefix and forecasts its
control-keypoint motion using four complementary experts: persistence, a
coefficient-space Transformer, a graph convolutional network, and a structured
Transformer with rigid and non-rigid motion constraints. A compact
routing table assigns expert mixtures across soft motion groups and
forecast-horizon bins, while a persistence gate suppresses unreliable
extrapolation.  Across D-NeRF and HyperNeRF scenes, MoRE-GS achieves  improved performance against other methods and these results demonstrate that renderer-calibrated composition
of complementary motion predictors improves causal dynamic Gaussian
forecasting across heterogeneous motion regimes.
\noindent\textbf{Project page:} \url{https://advanced-vision-and-learning-lab.github.io/MoRE-GS-website/}
\end{abstract}
\section{Introduction}
\label{sec:introduction}

Forecasting future 3D scene states is important for predictive perception and
planning in autonomous driving, closed-loop evaluation of safety-critical
autonomy systems, and model-based robotic navigation, where an agent must
anticipate scene evolution before future observations become
available~\cite{hu2021fiery,yang2023unisim,zheng2024occworld,
kayalibay2022spatial}. This problem is challenging because motion in dynamic scenes is both spatially
heterogeneous and temporally varying. Background regions may remain stationary,
objects may follow smooth or periodic trajectories, and articulated or
deformable components may exhibit locally coupled
motion~\cite{pumarola2021dnerf,ost2021neural}. Prediction reliability also
changes with the forecast horizon, defined as the temporal distance from
the last observed frame: recent motion can support accurate near-term
extrapolation, whereas trajectory, phase, and multimodal uncertainty accumulate
farther into the future~\cite{hu2021fiery}. Accurate future synthesis must
therefore adapt the motion model to both spatial region and forecast horizon.

3D Gaussian Splatting (3DGS)~\cite{kerbl2023gaussians} provides an explicit,
differentiable, and efficiently rendered representation for novel-view
synthesis. Dynamic extensions have achieved high-quality reconstruction by
optimizing Gaussian geometry, appearance, and deformation over observed
timestamps~\cite{luiten2024dynamic3dgs,wu2024fourdimensional,
yang2024deformable,li2024spacetime}. Future novel-view synthesis presents a
fundamentally different problem: the model must extrapolate open-loop to
timestamps beyond the observation interval while rendering from novel camera
views. Moreover, the complete future suffix must remain unavailable during
scene reconstruction, motion estimation, model selection, checkpoint
selection, and early stopping.

Existing forecasting approaches model motion using velocity fields, keypoint
graphs, continuous latent dynamics, or coherent motion
groups~\cite{li2023nvfi,zhao2024gaussianprediction,wang2026odegs,
Lee_2026_CVPR}. Each formulation provides a useful inductive bias, but existing
methods typically rely on a single forecasting family across the scene or
apply the same predictor family to all motion groups. A predictor that is
well suited to smooth or periodic trajectories may be less appropriate for
articulated, interacting, or nearly static regions. Moreover, the most
appropriate predictor may change with the forecasting horizon. This motivates
routing among complementary motion models at both the motion-group and
forecast-horizon levels.

We propose Motion Regime Expert Routing for Gaussian Splatting
(MoRE-GS), a causal, render-aware framework that freezes an observed-prefix
3DGS and routes complementary control-keypoint forecasts by motion group and
horizon. Its renderer-calibrated router includes a persistence gate
that suppresses unreliable extrapolation, and all routing parameters are learned from chronological partitions
within the observed prefix before future evaluation.

Our contributions are:
\begin{itemize}
    \item A heterogeneous motion-expert framework for forecasting a frozen
    dynamic 3D Gaussian representation by combining persistence, an
    orthonormal coefficient-space predictor, graph-based interactions, and
    rigidity-aware structured motion prediction.

    \item A render-aware, group- and horizon-conditioned router that separately
    controls persistence and combines complementary learned motion predictors
    through image-space supervision from the frozen Gaussian renderer.

    \item An observed-only temporal fit-calibrate-refit protocol that calibrates the router
    without exposing reconstruction, expert training, routing, checkpoint
    selection, or early stopping to the withheld future suffix.
\end{itemize}
\section{Related Work}
\label{sec:related_work}

\subsection{Static and dynamic scene representations}

Neural Radiance Fields (NeRF)~\cite{mildenhall2020nerf} use continuous
scene functions, whereas 3DGS uses explicit Gaussian primitives and
differentiable splatting. Under limited observations, both benefit from
geometric, appearance, or frequency-domain priors
~\cite{salehi2026lgdwtgs,zhu2024fsgs}. Dynamic extensions model temporal
change~\cite{pumarola2021dnerf,park2021hypernerf,luiten2024dynamic3dgs,
wu2024fourdimensional,yang2024deformable,li2024spacetime}, but primarily target
reconstruction or interpolation at observed times. MoRE-GS instead combines
complementary motion biases for causal extrapolation to a withheld suffix.


\subsection{Future dynamic scene forecasting}

Neural Velocity Fields for 3D Physics Learning from Dynamic Videos
(NVFi)~\cite{li2023nvfi} combines a radiance representation with a learned
velocity field to support temporal extrapolation.
GaussianPrediction~\cite{zhao2024gaussianprediction} introduces
lifecycle-aware canonical Gaussians and concentric motion distillation,
reducing dense Gaussian deformation to a compact keypoint graph. Future
keypoint translations and rotations are then predicted using a graph
convolutional network. MoRE-GS retains this Gaussian representation and
keypoint-based motion interface as a controlled foundation, while extending
the forecasting stage with complementary motion experts and scene-calibrated group-horizon routing.

Latent Ordinary Differential Equations for Dynamic Scene Extrapolation with 3D Gaussian Splatting (ODE-GS)~\cite{wang2026odegs} evolves Transformer-encoded Gaussian trajectories using a neural ordinary differential equation. Motion Group-aware Gaussian Forecasting (MoGaF)~\cite{Lee_2026_CVPR} groups Gaussians by motion and applies different constraints to approximately rigid and non-rigid regions. Unlike these approaches, MoRE-GS explicitly combines heterogeneous motion predictors through group- and horizon-dependent routing within a frozen Gaussian scene.

\subsection{Motion structure and expert integration}

DynMF~\cite{kratimenos2024dynmf} models motion through shared neural trajectories, while MoSca~\cite{lei2025mosca} uses a sparse 4D scaffold. MoE-GS~\cite{jin2026moegs} combines deformation experts through a volume-aware pixel router for observed-time reconstruction. In contrast, MoRE-GS routes heterogeneous control-keypoint predictors across 3D motion groups and forecast horizons for causal future extrapolation, with router calibration restricted to a disjoint interval within the observed prefix.
\section{Method}
\label{sec:method}

\begin{figure*}[t]
    \centering
    \includegraphics[width=\textwidth]{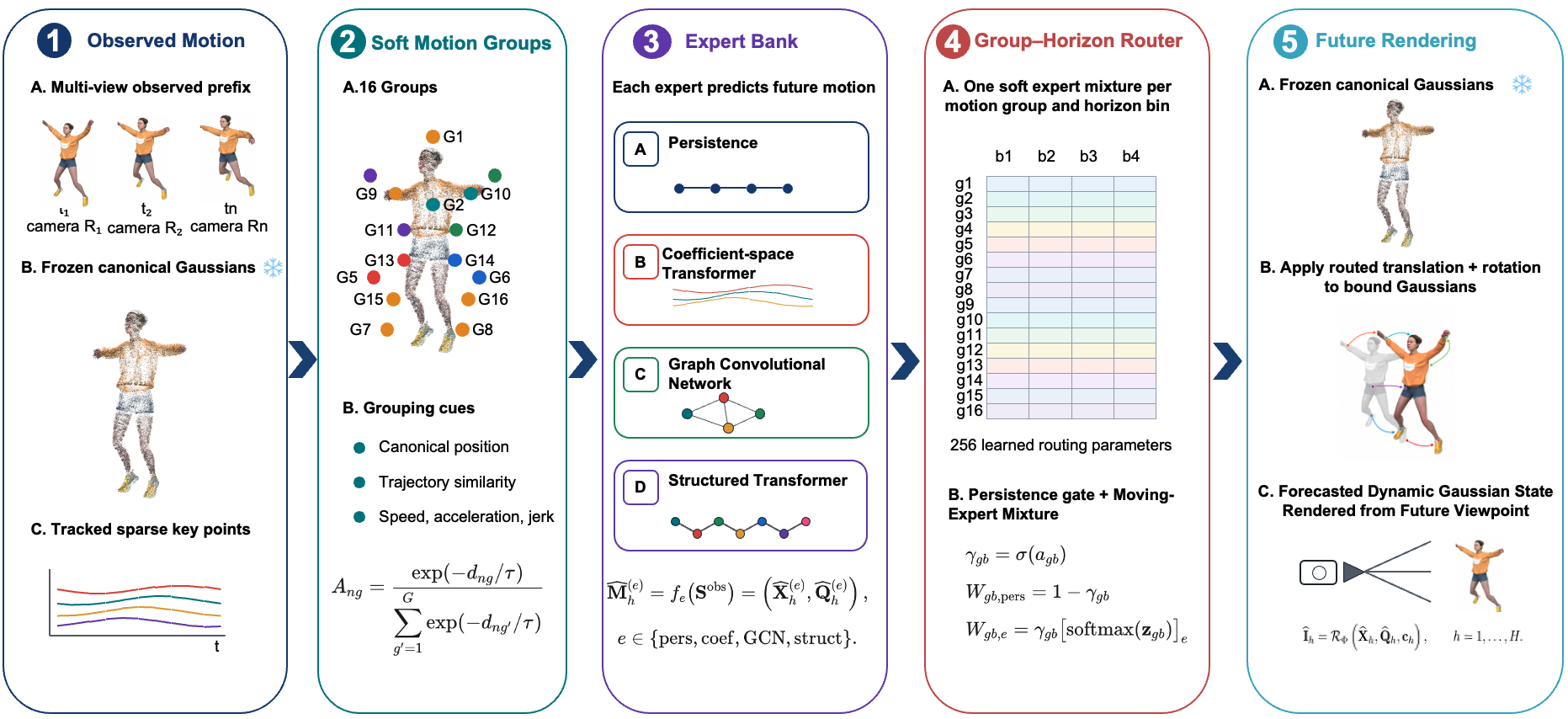}
    \caption{MoRE-GS forecasts control-keypoint motion using four complementary
    experts and an observed-only group--horizon router.}
    \label{fig:moregs_method}
\end{figure*}

\subsection{Problem Setup and Overview}
\label{sec:problem_setup}

Using only observed-prefix frames, we reconstruct the scene with the dynamic 3D Gaussian representation and renderer of GaussianPrediction~\cite{zhao2024gaussianprediction}. We then freeze Gaussian geometry, appearance, scale, opacity, deformation representation, control keypoints, keypoint-to-Gaussian motion bindings, and renderer. MoRE-GS forecasts future motion without reoptimizing the reconstruction.

Let frames $1,\ldots,T_{\mathrm{obs}}$ form the observed prefix and the
following $H$ frames form the withheld future suffix. GaussianPrediction
represents deformation using $N$ control keypoints with translation and
rotation trajectories. Given the final $L=10$ observed keypoint poses,
MoRE-GS predicts their transformations for the next $H$ steps. The frozen
bindings transfer these transformations to the Gaussians, and
Eq.~\ref{eq:future_render} renders the corresponding future view:
\begin{equation}
    \widehat{\mathbf I}_{h}
    =
    \mathcal{R}_{\Phi}
    \left(
        \widehat{\mathbf X}_{h},
        \widehat{\mathbf Q}_{h},
        \mathbf c_h
    \right),
    \qquad h=1,\ldots,H.
    \label{eq:future_render}
\end{equation}
Here, $\widehat{\mathbf I}_{h}$ is the predicted future image at forecast step
$h$, $\widehat{\mathbf X}_{h}\in\mathbb{R}^{N\times3}$ and
$\widehat{\mathbf Q}_{h}\in\mathbb{R}^{N\times4}$ contain the predicted
translations and quaternion rotations of the control keypoints,
$\mathbf c_h$ contains the evaluation-camera parameters, $\Phi$ denotes the
frozen Gaussian scene and motion bindings, and $\mathcal{R}_{\Phi}$ is the
frozen renderer.

Figure~\ref{fig:moregs_method} illustrates four heterogeneous motion experts combined by a scene-specific router across soft motion groups and forecast-horizon bins. A persistence gate suppresses unreliable extrapolation. The router is calibrated through the frozen renderer on a chronologically held-out portion of the observed prefix and frozen before future evaluation.

\subsection{Motion Experts}
\label{sec:motion_experts}

MoRE-GS uses four complementary experts to forecast control-keypoint translation and rotation from observed trajectories: persistence, a coefficient-space Transformer, a graph convolutional network (GCN), and a structured Transformer. All learned experts use observed trajectories only, with consistent training and output conventions across routing comparisons.

\paragraph{Persistence.}
The parameter-free persistence expert repeats the final observed translation and rotation at every future step, providing a conservative static-motion hypothesis for stationary regions or unreliable extrapolation.

\paragraph{Coefficient Transformer.}
Let $\widetilde{\mathbf S}_n\in\mathbb{R}^{L\times7}$ denote the channel-standardized motion history of keypoint $n$, comprising three
translation coordinates and four quaternion components. Quaternions are first
unit-normalized and sign-aligned; each channel is then standardized over the
$L$ observed steps using its temporal mean and standard deviation. Equation
\ref{eq:coefficient_projection} reparameterizes this history using a fixed,
seeded random orthonormal basis:
\begin{equation}
    \mathbf Z_n
    =
    \mathbf B\widetilde{\mathbf S}_n,
    \qquad
    \mathbf B\in\mathbb{R}^{L\times L},
    \qquad
    \mathbf B^\top\mathbf B=\mathbf I,
    \label{eq:coefficient_projection}
\end{equation}
where $\mathbf Z_n\in\mathbb{R}^{L\times7}$ contains the transformed
coefficients, $\mathbf B$ is the temporal basis, and $\mathbf I$ is the
$L\times L$ identity matrix. Orthonormality preserves trajectory norms and
pairwise distances:
$\|\mathbf B\widetilde{\mathbf S}_n\|_F
=\|\widetilde{\mathbf S}_n\|_F$.
Since $\mathbf B$ acts only along time, it never mixes translation and
quaternion channels. With all coefficients retained, its exact inverse
$\mathbf B^\top$ and unit condition number give lossless, scale-neutral
temporal tokens.

The coefficients are partitioned into three blocks and combined with
canonical-position and forecast-horizon embeddings. A shared one-layer
Transformer encoder~\cite{vaswani2017attention} and gated decoder predict
translation and rotation residuals relative to persistence for all future
steps in parallel. Each seven-dimensional coefficient is embedded at
width 32 with coefficient-index, block, and canonical-position embeddings.
The encoder pools three block contexts; a horizon-conditioned gated decoder
returns $N\times H\times7$ residuals relative to persistence. Translation
residuals are rescaled and added, while quaternions are renormalized. Pose,
velocity, acceleration, and gate losses are specified in the supplement. The random basis does not impose a predetermined frequency
ordering. It outperforms the frequency-ordered discrete cosine transform type
II (DCT-II)~\cite{ahmed1974dct} in the supplementary matched ablation;
therefore, we assign no physical frequency interpretation to its coefficient
blocks.

\paragraph{Graph expert.}
The GCN uses the forecasting architecture released with
GaussianPrediction~\cite{zhao2024gaussianprediction}. Separate translation and
rotation branches learn affinities among control keypoints and propagate
coordinated motion across related regions. It provides a relational motion
hypothesis that complements the parallel temporal prediction of the
coefficient Transformer. The GCN predicts one future step at a time and is
rolled out autoregressively over the forecast interval.

\paragraph{Structured expert and hard groups.}
The structured Transformer captures group-level dynamics within coherent
keypoint groups. Its motion-aware grouping is motivated by
MoGaF~\cite{Lee_2026_CVPR}, while its architecture and supervision are specific
to MoRE-GS. Each keypoint descriptor concatenates its canonical position with
the first six principal component analysis (PCA) coefficients of its observed
translation trajectory. We use the same empirically selected rank of six for
every scene. In an observed-only analysis, these components retain $99.87\%$
of the displacement variance on average across the eight D-NeRF scenes,
supporting a compact representation with negligible loss of observed motion
information.

The descriptors are divided into eight hard groups, and a one-layer masked
Transformer models each group. Within each group, a
shared one-layer Transformer projects each standardized $L\times7$ keypoint
history to width 32 and applies temporal self-attention within that keypoint.
A contiguous masked span is annealed during training; a linear head predicts
the next pose, and autoregressive rollout produces the future sequence. A normalized Kabsch alignment
residual~\cite{kabsch1976solution} classifies each group as approximately rigid
or non-rigid. We align its first observed keypoint configuration with sampled
subsequent configurations and normalize the root-mean-square alignment
residual by the root-mean-square group radius. A group is classified as rigid
when the temporal-median normalized residual is at most $0.10$ and as
non-rigid otherwise. Rigid groups receive a radius-normalized Kabsch alignment
loss, whereas non-rigid groups receive a radius-normalized local
displacement-consistency loss over four nearest-neighbor edges. These
physical terms are combined with pose and acceleration losses (weights
are reported in the supplement). These hard
groups are internal to the structured expert and are distinct from the soft
router groups defined below.

\subsection{Soft Groups and Group-Horizon Router}
\label{sec:router}

\paragraph{Soft router groups.}
We construct $G=16$ soft groups before router calibration. Canonical
positions come from the frozen reconstruction, while trajectory-derived
descriptor terms use only the expert-fitting interval. Each keypoint descriptor contains its canonical position, up
to eight uniformly sampled scene-normalized displacements, and observed speed,
acceleration, and jerk statistics. After standardization, we apply soft
$k$-means with deterministic farthest-point initialization: the first center
is the descriptor with the largest squared norm, subsequent centers maximize
distance to the nearest existing center, and the centers are updated for 30
fixed iterations without random sampling.

The resulting assignment matrix
$\mathbf A\in[0,1]^{N\times G}$ is defined by
Eq.~\ref{eq:soft_assignment}:
\begin{equation}
    A_{ng}
    =
    \frac{\exp(-d_{ng}/\tau)}
         {\sum_{g'=1}^{G}\exp(-d_{ng'}/\tau)},
    \label{eq:soft_assignment}
\end{equation}
where $A_{ng}$ is the membership of keypoint $n$ in group $g$, $d_{ng}$ is
its normalized squared distance to group center $g$, and $\tau$ controls
assignment softness. The assignment matrix is fixed before router calibration,
allowing boundary keypoints to share routing weights across compatible groups.

\paragraph{Routing weights.}
We divide the forecast interval into $B=4$ contiguous bins and let $b(h)$
denote the bin containing forecast step $h$. Bins use relative progress,
$b(h)=1+\min\{B-1,\lfloor(h-1)B/H\rfloor\}$.The moving expert set and complete
expert set are
\[
\mathcal E_{\mathrm{mov}}
=\{\mathrm{coef},\mathrm{GCN},\mathrm{struct}\},
\qquad
\mathcal E
=\mathcal E_{\mathrm{mov}}\cup\{\mathrm{pers}\}.
\]
For every group $g$ and horizon bin $b$, Eq.~\ref{eq:router_weights} defines a
persistence gate and the moving-expert weights:
\begin{equation}
\begin{aligned}
    \gamma_{gb}
    &=
    \sigma(a_{gb}),\\
    W_{gb,e}
    &=
    \begin{cases}
        1-\gamma_{gb},
        & e=\mathrm{pers},\\[2pt]
        \gamma_{gb}
        \left[
            \operatorname{softmax}(\mathbf z_{gb})
        \right]_e,
        & e\in\mathcal E_{\mathrm{mov}}.
    \end{cases}
\end{aligned}
\label{eq:router_weights}
\end{equation}
Here, $a_{gb}$ is the scalar persistence-gate logit, $\sigma$ is the sigmoid
function, $\gamma_{gb}\in[0,1]$ is the total probability assigned to moving
experts, $\mathbf z_{gb}\in\mathbb{R}^{3}$ contains their logits, and
$W_{gb,e}$ is the weight assigned to expert $e$. The gate determines how much
motion is used, while the softmax distributes this probability among the
three learned experts. With 16 groups and 4 bins, the router contains
$16\times4\times4=256$ trainable parameters and is a scene-specific calibrated
table rather than an input-conditioned network.

Equation~\ref{eq:keypoint_weights} converts group-level weights into the
expert weight for keypoint $n$ at horizon $h$:
\begin{equation}
    \pi_{nhe}
    =
    \sum_{g=1}^{G}
    A_{ng}W_{g,b(h),e},
    \label{eq:keypoint_weights}
\end{equation}
where $\pi_{nhe}$ is the final weight assigned to expert $e$.

The routed translation is defined in Eq.~\ref{eq:translation_mixture}:
\begin{equation}
    \widehat{\mathbf x}_{n,h}
    =
    \sum_{e\in\mathcal E}
    \pi_{nhe}\widehat{\mathbf x}^{e}_{n,h},
    \label{eq:translation_mixture}
\end{equation}
where $\widehat{\mathbf x}^{e}_{n,h}\in\mathbb{R}^{3}$ is expert $e$'s
candidate translation and $\widehat{\mathbf x}_{n,h}$ is the routed result.

Because a quaternion and its negative represent the same rotation,
Eq.~\ref{eq:rotation_mixture} aligns each candidate with the persistence
quaternion before computing a normalized weighted average:
\begin{align}
    \widetilde{\mathbf q}^{e}_{n,h}
    &=
    \operatorname{sign}
    \left(
        \left\langle
        \widehat{\mathbf q}^{e}_{n,h},
        \widehat{\mathbf q}^{\mathrm{pers}}_{n,h}
        \right\rangle
    \right)
    \widehat{\mathbf q}^{e}_{n,h},
    \notag\\
    \widehat{\mathbf q}_{n,h}
    &=
    \frac{
        \sum_{e\in\mathcal E}
        \pi_{nhe}\widetilde{\mathbf q}^{e}_{n,h}
    }{
        \left\|
        \sum_{e\in\mathcal E}
        \pi_{nhe}\widetilde{\mathbf q}^{e}_{n,h}
        \right\|_2
    }.
    \label{eq:rotation_mixture}
\end{align}
Here, $\widehat{\mathbf q}^{e}_{n,h}$ is expert $e$'s candidate quaternion,
$\widehat{\mathbf q}^{\mathrm{pers}}_{n,h}$ is the persistence quaternion,
$\widetilde{\mathbf q}^{e}_{n,h}$ is its sign-aligned form, and
$\widehat{\mathbf q}_{n,h}$ is the final unit quaternion.

\subsection{Observed-Only Router Calibration}
\label{sec:router_calibration}

Calibrating the router on the future suffix would introduce evaluation
information into model selection. We therefore use an observed-only temporal fit—calibrate—refit procedure
with a fixed chronological 75/25 division of the observed prefix. The first
75\% is used to fit temporary experts, while the remaining 25\% forms a
contiguous calibration interval excluded from temporary-expert fitting.
It is included in the frozen reconstruction, so ``out-of-fit'' applies
only to the temporary motion experts, not to representation learning. Under the primary
D-NeRF 80/20 protocol, this calibration interval nominally matches the
duration of the sealed future suffix, up to rounding. The division is fixed
across scenes and is not selected using future performance.

The temporary
experts are then discarded. Independently initialized deployment experts with
the same architectures, preprocessing, objectives, and output conventions are
trained on the complete observed prefix and paired with the frozen router. No
future evaluation frame, motion state, image, or metric is used for
reconstruction, grouping, expert training, router calibration, checkpoint
selection, or early stopping. Relative-progress bins preserve bin meaning
across calibration and deployment. Refit experts retain the same normalization,
architecture, losses, and output interface, so the router mixes matched
quantities. This transfer is empirical; split robustness is reported in the
supplement.

\subsection{Render-Aware Router Objective}
\label{sec:render_loss}

Keypoint-space error weights visible and occluded motion equally, although
their effects on the rendered image may differ substantially. We therefore
calibrate the router using image-space error from the frozen renderer. For
pixel $p$ in calibration image $\mathbf I_h$, Eq.~\ref{eq:pixel_scores}
defines foreground and dynamic-change scores:
\begin{align}
    \delta^{\mathrm{fg}}_h(p)
    &=
    \frac{1}{3}
    \left\|
        \mathbf I_h(p)-\mathbf I_{\mathrm{bg}}
    \right\|_1,
    \notag\\
    \delta^{\mathrm{dyn}}_h(p)
    &=
    \frac{1}{3}
    \left\|
        \mathbf I_h(p)
        -
        \widehat{\mathbf I}^{\mathrm{pers}}_h(p)
    \right\|_1.
    \label{eq:pixel_scores}
\end{align}
Here, $\delta^{\mathrm{fg}}_h(p)$ measures mean absolute RGB difference from
the background, $\delta^{\mathrm{dyn}}_h(p)$ measures mean absolute RGB
difference from persistence, $\mathbf I_{\mathrm{bg}}$ is the background
color, and $\widehat{\mathbf I}^{\mathrm{pers}}_h$ is the persistence
rendering.

Equation~\ref{eq:importance_mask} defines the important-pixel mask:
\begin{equation}
    \mathcal M_h
    =
    \left\{
        p:
        \delta^{\mathrm{fg}}_h(p)>\tau_{\mathrm{fg}}
        \ \lor\
        \delta^{\mathrm{dyn}}_h(p)>\tau_{\mathrm{dyn}}
    \right\},
    \label{eq:importance_mask}
\end{equation}
where $\tau_{\mathrm{fg}}$ and $\tau_{\mathrm{dyn}}$ are fixed foreground and
dynamic-change thresholds. This mask emphasizes visible content and regions
not explained by persistence.

For calibration frame $h$, Eq.~\ref{eq:rendering_objective} defines the
rendering loss:
\begin{equation}
    \ell^{(h)}_{\mathrm{render}}
    =
    \operatorname{MSE}
    \left(
        \widehat{\mathbf I}_h,
        \mathbf I_h
    \right)
    +
    \operatorname{MSE}_{p\in\mathcal M_h}
    \left(
        \widehat{\mathbf I}_h,
        \mathbf I_h
    \right),
    \label{eq:rendering_objective}
\end{equation}
where the first mean-squared error (MSE) term is evaluated over the complete
image and the second over pixels in $\mathcal M_h$. The aggregate rendering
loss is
\begin{equation}
    \mathcal L_{\mathrm{render}}
    =
    \frac{1}{T_{\mathrm{cal}}}
    \sum_{h=1}^{T_{\mathrm{cal}}}
    \ell^{(h)}_{\mathrm{render}}.
    \label{eq:rendering_aggregate}
\end{equation}

The router uses three regularizers:
\begin{align}
    \mathcal L_{\mathrm{smooth}}
    &=
    \frac{1}{G(B-1)|\mathcal E|}
    \sum_{g=1}^{G}
    \sum_{b=1}^{B-1}
    \sum_{e\in\mathcal E}
    \left(
        W_{g,b+1,e}-W_{g,b,e}
    \right)^2,
    \label{eq:router_smoothness}\\
    \mathcal L_{\mathrm{move}}
    &=
    \frac{1}{GB}
    \sum_{g=1}^{G}
    \sum_{b=1}^{B}
    \sum_{e\in\mathcal E_{\mathrm{mov}}}
    W_{gb,e}
    =
    \frac{1}{GB}
    \sum_{g=1}^{G}
    \sum_{b=1}^{B}
    \gamma_{gb},
    \label{eq:router_moving_mass}\\
    \mathcal L_{\mathrm{entropy}}
    &=
    -\frac{1}{GB}
    \sum_{g=1}^{G}
    \sum_{b=1}^{B}
    \sum_{e\in\mathcal E_{\mathrm{mov}}}
    W_{gb,e}
    \log\!\left(\max(W_{gb,e},\epsilon)\right).
    \label{eq:router_entropy}
\end{align}
Here, $|\mathcal E|$ is the number of experts and $\epsilon=10^{-8}$ prevents
numerical instability. The smoothness term discourages abrupt routing changes
between adjacent horizon bins, the moving-mass term favors persistence unless
motion reduces the rendered error, and the entropy term encourages
concentrated moving-expert weights.

The complete router-calibration objective is
\begin{equation}
    \mathcal L_{\mathrm{router}}
    =
    \mathcal L_{\mathrm{render}}
    +
    \lambda_{\mathrm{s}}\mathcal L_{\mathrm{smooth}}
    +
    \lambda_{\mathrm{m}}\mathcal L_{\mathrm{move}}
    +
    \lambda_{\mathrm{e}}\mathcal L_{\mathrm{entropy}},
    \label{eq:router_objective}
\end{equation}
where $\lambda_{\mathrm{s}}$, $\lambda_{\mathrm{m}}$, and
$\lambda_{\mathrm{e}}$ are the nonnegative regularization weights. Exact mask
thresholds, regularization weights, and optimization hyperparameters are
provided in the supplementary material.

\section{Experiments}
\label{sec:experiments}

\subsection{Experimental Setup}
\label{sec:setup}

We evaluate on all eight D-NeRF scenes~\cite{pumarola2021dnerf} at
$800\!\times\!800$ resolution. Each training-view sequence is ordered
chronologically: the first 80\% is observed, and the final 20\% is reserved
exclusively for forecasting evaluation. The Gaussian representation is
reconstructed for 60K iterations using only the observed prefix. HyperNeRF
~\cite{park2021hypernerf} provides a secondary real-capture stress test. We
retain only the official training-camera sequence, exclude the held-out
validation cameras, and apply a chronological 90/10 observed/future split at
the native scene resolution using the released scene-specific reconstruction
configuration. Thus, future images are never used for representation fitting,
grouping, expert training, router calibration, checkpoint selection, or early
stopping.

\paragraph{Forecasting configuration.}
The observed-only Gaussian representation is frozen before forecasting. Each
expert receives the final ten keypoint poses. Both Transformers use one layer,
width 32, eight heads, feed-forward width 64, and dropout 0.2, and are trained
for 2,001 updates with learning rate $10^{-3}$ to $10^{-5}$; the GCN follows
GaussianPrediction. MoRE-GS uses 16 groups, four horizon bins, and a
256-parameter router. Temporary experts fit the first 75\% of the observed
prefix and calibrate the router on the remaining 25\% for 200 AdamW updates
at learning rate 0.05. They are then replaced by deployment experts trained on
the full observed prefix, while the router remains frozen.

We report peak signal-to-noise ratio (PSNR), structural similarity index
measure (SSIM), and learned perceptual image patch similarity (LPIPS). We first
average over evaluation views within each scene and then weight all scenes
equally. Every reported checkpoint is cold-reloaded before evaluation. Forecasting-head stability over seeds 17, 42, and 123 is reported in the
supplementary material.
GaussianPrediction and ODE-GS results are obtained from our local
reproductions, whereas MoGaF values are paper-reported system-level
comparisons and are not treated as matched ablations.

\subsection{D-NeRF Forecasting}
\label{sec:dnerf-results}
Table~\ref{tab:dnerf_main} shows that MoRE-GS achieves the highest aggregate
performance, with 27.47 dB PSNR, 0.9480 SSIM, and 0.0446 LPIPS. It improves
mean PSNR by 2.13 dB over GaussianPrediction, 1.60 dB over MoGaF, and
0.82 dB over ODE-GS. MoRE-GS achieves the highest PSNR on four of the eight
scenes and ranks first in 16 of the 24 per-scene metric comparisons, including
all three metrics on Mutant, Bouncing Balls, and Lego. On T-Rex, Jumping
Jacks, and Hook, ODE-GS obtains higher PSNR, whereas MoRE-GS achieves higher
SSIM and lower LPIPS. These results demonstrate strong aggregate performance
and complementary improvements in pixel, structural, and perceptual quality.
Hell Warrior remains the principal failure case: MoRE-GS trails
ODE-GS in PSNR/SSIM and MoGaF in all three metrics.

\begin{table*}[t]
\centering
\caption{D-NeRF 80/20 future forecasting. Best values, determined from the
unrounded scores, are shown in bold.}
\label{tab:dnerf_main}
\fontsize{6.5}{7.4}\selectfont
\setlength{\tabcolsep}{1.5pt}
\renewcommand{\arraystretch}{1.12}

\begin{tabular*}{\textwidth}{
@{\extracolsep{\fill}}l*{12}{c}@{}}
\toprule
& \multicolumn{3}{c}{GaussianPrediction}
& \multicolumn{3}{c}{MoGaF}
& \multicolumn{3}{c}{ODE-GS}
& \multicolumn{3}{c}{MoRE-GS} \\
\cmidrule(lr){2-4}
\cmidrule(lr){5-7}
\cmidrule(lr){8-10}
\cmidrule(lr){11-13}
Scene
& PSNR $\uparrow$ & SSIM $\uparrow$ & LPIPS $\downarrow$
& PSNR $\uparrow$ & SSIM $\uparrow$ & LPIPS $\downarrow$
& PSNR $\uparrow$ & SSIM $\uparrow$ & LPIPS $\downarrow$
& PSNR $\uparrow$ & SSIM $\uparrow$ & LPIPS $\downarrow$ \\
\midrule

T-Rex
& 21.08 & 0.940 & \textbf{0.046}
& 20.60 & 0.941 & 0.050
& \textbf{21.51} & 0.944 & 0.054
& 21.13 & \textbf{0.944} & 0.048 \\

Mutant
& 27.55 & 0.953 & 0.027
& 28.72 & 0.955 & 0.022
& 31.01 & 0.967 & 0.026
& \textbf{33.55} & \textbf{0.976} & \textbf{0.015} \\

Jumping Jacks
& 20.00 & 0.912 & 0.084
& 20.12 & 0.922 & 0.073
& \textbf{22.23} & 0.920 & 0.080
& 21.18 & \textbf{0.925} & \textbf{0.069} \\

Stand Up
& 27.44 & 0.948 & 0.044
& 30.38 & 0.954 & \textbf{0.034}
& 27.23 & 0.944 & 0.054
& \textbf{30.68} & \textbf{0.958} & 0.034 \\

Bouncing Balls
& 27.09 & 0.972 & 0.030
& 26.82 & 0.973 & 0.037
& 25.76 & 0.964 & 0.053
& \textbf{28.40} & \textbf{0.977} & \textbf{0.030} \\

Hook
& 24.21 & 0.914 & 0.054
& 24.13 & 0.908 & 0.056
& \textbf{26.99} & 0.934 & 0.054
& 26.47 & \textbf{0.937} & \textbf{0.041} \\

Hell Warrior
& 31.03 & 0.929 & 0.073
& 31.29 & \textbf{0.932} & \textbf{0.068}
& \textbf{31.66} & 0.931 & 0.083
& 30.22 & 0.918 & 0.081 \\

Lego
& 24.30 & 0.919 & 0.063
& 24.88 & 0.901 & 0.053
& 26.75 & 0.935 & 0.053
& \textbf{28.10} & \textbf{0.949} & \textbf{0.039} \\

\midrule
Mean
& 25.34 & 0.936 & 0.053
& 25.87 & 0.936 & 0.049
& 26.64 & 0.942 & 0.057
& \textbf{27.47} & \textbf{0.948} & \textbf{0.045} \\
\bottomrule
\end{tabular*}
\end{table*}

\begin{figure*}[t]
  \centering
  \includegraphics[width=\textwidth]{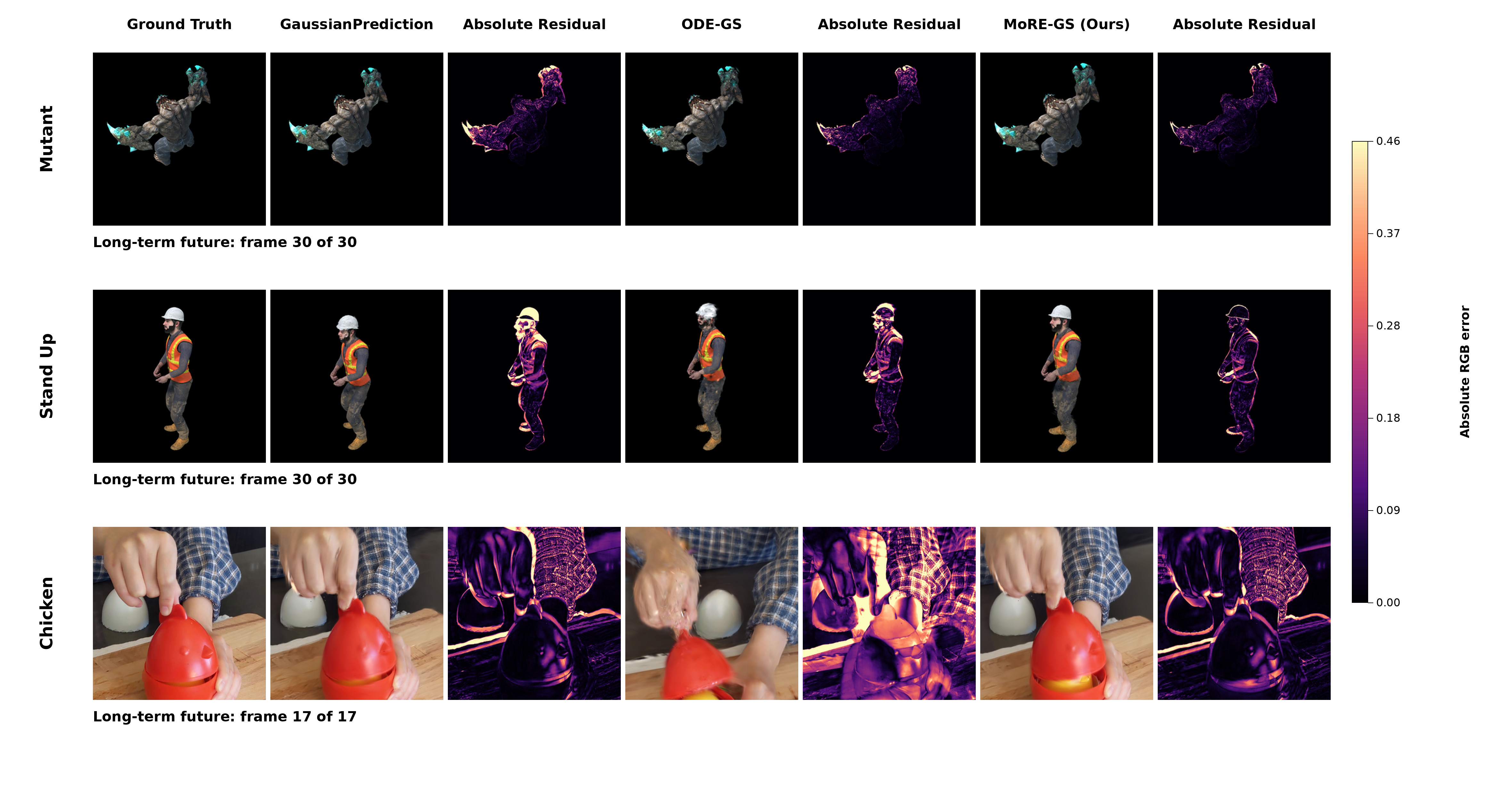}
  \caption{Long-horizon forecasts and absolute RGB residuals on Mutant,
  Stand Up, and Chicken. All residual maps use the same color scale.}
  \label{fig:long-horizon-residuals}
\end{figure*}

Figure~\ref{fig:long-horizon-residuals} provides qualitative evidence at the
longest forecast horizon. On Mutant and Stand Up, MoRE-GS better preserves
global pose and reduces foreground residuals. On the real-capture Chicken
scene, it substantially reduces the motion distortion produced by ODE-GS.

Across D-NeRF, MoRE-GS improves mean PSNR over persistence by 3.124, 3.169,
3.021, and 1.728 dB from the nearest to the farthest future quartile. The
1.728 dB improvement at the farthest horizon confirms that its advantage is
not limited to near-term prediction. Complete quartile-level results are
provided in the supplementary material.

\subsection{HyperNeRF Results}
\label{sec:hypernerf-results}

HyperNeRF provides a challenging real-capture evaluation involving handheld
cameras, occlusion, non-rigid interactions, and appearance variation.
Table~\ref{tab:hypernerf} reports results under the causal chronological
90/10 observed/future protocol using 16 motion groups. MoRE-GS improves mean
PSNR by 0.389 dB and reduces mean LPIPS by 0.0496 relative to ODE-GS
reproduction. It achieves higher PSNR on three of four scenes and lower LPIPS
on all four scenes. ODE-GS retains a 0.0121 advantage in mean SSIM, indicating
that motion-only expert routing does not fully preserve fine appearance and
visibility structure under real capture.

\begin{figure*}[t]
  \centering
  \includegraphics[width=0.96\textwidth]{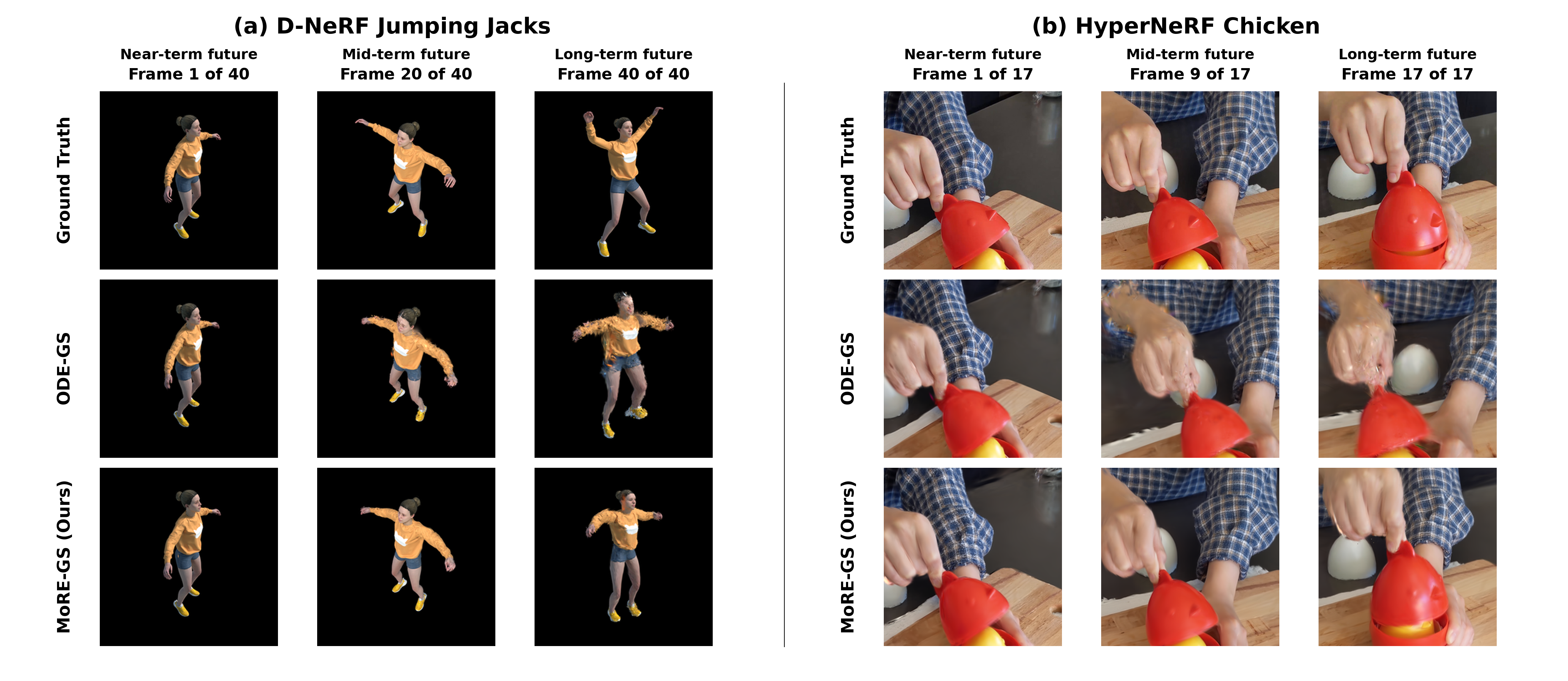}
  \caption{Near-, middle-, and long-horizon forecasts on D-NeRF Jumping Jacks
  and HyperNeRF Chicken.}
  \label{fig:qualitative-horizons}
\end{figure*}

Figure~\ref{fig:qualitative-horizons} compares near-, middle-, and long-horizon
predictions on D-NeRF Jumping Jacks and HyperNeRF Chicken. MoRE-GS better
preserves the dominant pose and group-level trajectory because its
group--horizon router can favor different motion predictors as extrapolation
difficulty changes. Residual errors concentrate near articulated boundaries
and occlusions, where small keypoint or rotation errors produce large
pixel-space misalignment. Errors are more pronounced in appearance-changing
regions because MoRE-GS forecasts only translation and rotation while keeping
geometry, appearance, and opacity frozen. It therefore cannot create newly
visible content or model temporal appearance changes. These results indicate
that routing primarily improves motion extrapolation, while appearance and
visibility forecasting remain important limitations.
Additional examples at near, middle, and long forecast horizons are provided in the
supplementary material.

\begin{table}[t]
\centering
\caption{HyperNeRF future forecasting under the chronological 90/10 protocol.
MoRE-GS uses 16 motion groups. Best values are shown in bold.}
\label{tab:hypernerf}
\scriptsize
\setlength{\tabcolsep}{3.2pt}
\renewcommand{\arraystretch}{1.10}

\resizebox{\columnwidth}{!}{%
\begin{tabular}{@{}l*{6}{c}@{}}
\toprule
&
\multicolumn{3}{c}{ODE-GS}
&
\multicolumn{3}{c}{MoRE-GS}
\\
\cmidrule(lr){2-4}
\cmidrule(lr){5-7}

Scene
& PSNR$\uparrow$ & SSIM$\uparrow$ & LPIPS$\downarrow$
& PSNR$\uparrow$ & SSIM$\uparrow$ & LPIPS$\downarrow$
\\
\midrule

Chicken
& 18.49 & \textbf{0.594} & 0.467
& \textbf{19.52} & 0.581 & \textbf{0.368}
\\

3D Printer
& \textbf{16.73} & \textbf{0.543} & 0.485
& 16.45 & 0.519 & \textbf{0.423}
\\

Cut Lemon
& 20.46 & \textbf{0.617} & 0.379
& \textbf{20.97} & 0.598 & \textbf{0.345}
\\

Split Cookie
& 19.65 & 0.630 & 0.264
& \textbf{19.95} & \textbf{0.638} & \textbf{0.259}
\\
\midrule

Mean
& 18.83 & \textbf{0.596} & 0.399
& \textbf{19.22} & 0.584 & \textbf{0.349}
\\
\bottomrule
\end{tabular}%
}
\end{table}

\subsection{Causal Ablations}
\label{sec:causal-ablations}

Table~\ref{tab:core-ablation} evaluates expert choice, routing design, grouping,
group count, and regularization under matched settings. Individual-expert rows
use one predictor without composition. Fixed routing selects one expert on the
observed calibration interval, uniform routing equally mixes experts, and
scene-level routing learns one mixture for the entire scene. Horizon-only and
group-only routing vary the mixture across only forecast bins or motion groups,
respectively, whereas full MoRE-GS conditions on both. Additional controls
remove persistence, shuffle group assignments, vary group count, or remove all
router regularizers. All group-dependent controls use $G=16$ unless stated
otherwise.

MoRE-GS improves PSNR by 0.96 dB over the strongest individual expert,
0.86 dB over observed-only expert selection, and 0.81 dB over uniform mixing.
Removing persistence reduces PSNR by 0.62 dB, while motion-aware grouping
outperforms shuffled groups by 0.55 dB. These results attribute the primary
gain to heterogeneous expert composition, meaningful grouping, and
persistence abstention.

\begin{table}[t]
\centering
\caption{Causal D-NeRF ablations averaged over all eight scenes. Best overall
values, determined from unrounded results, are bold; the selected configuration
is underlined.}
\label{tab:core-ablation}
\fontsize{6.3}{7.0}\selectfont
\setlength{\tabcolsep}{2.2pt}
\renewcommand{\arraystretch}{1.06}
\begin{tabular}{@{}lccc@{}}
\toprule
Variant & PSNR $\uparrow$ & SSIM $\uparrow$ & LPIPS $\downarrow$ \\
\midrule
\multicolumn{4}{@{}l}{Individual experts} \\
Persistence
& 24.70 & 0.936 & 0.053 \\
Coefficient Transformer
& 26.51 & 0.941 & 0.051 \\
GCN
& 25.22 & 0.934 & 0.052 \\
Structured Transformer
& 25.95 & 0.932 & 0.061 \\
\midrule
\multicolumn{4}{@{}l}{Routing design} \\
Observed-only fixed expert
& 26.61 & 0.942 & 0.051 \\
Uniform mixture
& 26.66 & 0.944 & 0.045 \\
Scene-level routing
& 27.13 & 0.945 & 0.047 \\
Horizon-only routing
& 27.07 & 0.946 & 0.046 \\
Group-only routing
& 27.38 & 0.946 & 0.046 \\
Without persistence gate
& 26.85 & 0.944 & 0.048 \\
Shuffled groups
& 26.92 & 0.943 & 0.050 \\
\midrule
\multicolumn{4}{@{}l}{Group-count sensitivity} \\
MoRE-GS ($G=8$)
& 27.26 & 0.947 & 0.046 \\
\underline{MoRE-GS ($G=16$, selected)}
& \underline{27.47} & \underline{0.948} & \underline{0.045} \\
MoRE-GS ($G=32$)
& \textbf{27.66} & \textbf{0.949} & \textbf{0.044} \\
\midrule
\multicolumn{4}{@{}l}{Regularization} \\
MoRE-GS ($G=16$), no regularizers
& 27.34 & 0.948 & 0.045 \\
\bottomrule
\end{tabular}
\end{table}

Full group-horizon routing improves over group-only routing by 0.09 dB
PSNR and 0.002 SSIM, while reducing LPIPS by 0.002. Although the aggregate
gain is modest, the horizon-resolved analysis in the supplementary material
shows that the benefit is concentrated at middle-to-long forecast intervals.
These results support the two-dimensional routing design: motion groups capture
spatially heterogeneous dynamics, while horizon conditioning adapts expert
selection as extrapolation uncertainty changes.

Increasing $G$ from 8 to 16 improves PSNR by 0.20 dB, while $G=32$ provides
a further 0.20 dB improvement. We retain the cross-dataset setting $G=16$;
$G=32$ is reported only as a post-hoc D-NeRF sensitivity result. The selected
router contains only 256 parameters. Removing all regularizers reduces PSNR
by 0.13 dB and slightly degrades SSIM and LPIPS, confirming that regularization
provides a useful but secondary improvement.
\subsection{Efficiency}
\label{sec:efficiency}

Table~\ref{tab:efficiency} reports post-canonical efficiency averaged across
all eight D-NeRF scenes on one RTX 4090. MoRE-GS reduces training time from
10,444 s to 1,135 s, corresponding to a $9.20\times$ speedup over ODE-GS.
It also increases complete-pipeline throughput from 39.00 to 84.83 FPS
($2.18\times$) and reduces peak GPU memory by 11.5\%. These measurements
include evaluation of the complete expert bank and routing operation.

\begin{table}[t]
\centering
\caption{Post-canonical efficiency,
averaged over eight D-NeRF scenes. State excludes reconstructed scenes.}
\label{tab:efficiency}
\fontsize{6.1}{6.9}\selectfont
\setlength{\tabcolsep}{1.8pt}
\renewcommand{\arraystretch}{1.08}
\resizebox{\columnwidth}{!}{%
\begin{tabular}{lccccc}
\toprule
Method & Params & State & Train (s) & FPS & VRAM \\
\midrule
GaussianPrediction
& 14.173M & 69.26 MiB & -- & \textbf{155.11} & -- \\
ODE-GS
& \textbf{1.093M} & \textbf{13.10 MiB} & 10,444 & 39.00 & 2.43 GiB \\
MoRE-GS
& 14.280M & 69.75 MiB & \textbf{1,135} & 84.83 & \textbf{2.15 GiB} \\
\bottomrule
\end{tabular}%
}
\end{table}

Although MoRE-GS contains more forecasting parameters than ODE-GS, its shallow
experts permit substantially faster optimization and inference. Its additional
cost relative to GaussianPrediction arises from evaluating complementary
motion experts, while complete-pipeline throughput remains more than twice
that of ODE-GS.
\paragraph{Limitations.}
MoRE-GS demonstrates strong forecasting performance on both synthetic and real-world scenes. Extending its evaluation to more complex and unconstrained environments remains an important direction for future research. The current approach uses scene-specific router calibration, while cross-scene generalization offers opportunities for further investigation. Although multiple motion experts introduce additional computational overhead, the framework maintains favorable efficiency compared with ODE-GS under the evaluated settings. Finally, MoRE-GS focuses on forecasting motion within a fixed Gaussian representation, providing a foundation for future extensions that jointly model dynamic appearance, visibility, and scene evolution.
\section{Conclusion}
\label{sec:conclusion}

We presented MoRE-GS, a causal, render-aware framework that combines
complementary motion experts through group- and horizon-conditioned routing.
The router is calibrated entirely within the observed prefix, preventing
future information from influencing training or model selection. MoRE-GS achieves the best average performance on D-NeRF, improving PSNR by
0.82 dB over ODE-GS reproduction while providing substantially
faster post-canonical training and inference. Results on HyperNeRF further
demonstrate useful transfer to real captures. Overall, the results support
renderer-calibrated expert composition as an accurate and efficient approach
to future dynamic Gaussian forecasting.

{\small
\bibliographystyle{ieeenat_fullname}
\bibliography{references}

@article{kerbl2023gaussians,
  author  = {Kerbl, Bernhard and Kopanas, Georgios and Leimk{\"u}hler, Thomas and Drettakis, George},
  title   = {{3D Gaussian Splatting for Real-Time Radiance Field Rendering}},
  journal = {ACM Transactions on Graphics},
  volume  = {42},
  number  = {4},
  pages   = {139:1--139:14},
  year    = {2023},
  doi     = {10.1145/3592433}
}

@inproceedings{mildenhall2020nerf,
  author    = {Mildenhall, Ben and Srinivasan, Pratul P. and Tancik, Matthew and Barron, Jonathan T. and Ramamoorthi, Ravi and Ng, Ren},
  title     = {{NeRF}: Representing Scenes as Neural Radiance Fields for View Synthesis},
  booktitle = {European Conference on Computer Vision},
  pages     = {405--421},
  year      = {2020},
  doi       = {10.1007/978-3-030-58452-8_24}
}

@inproceedings{pumarola2021dnerf,
  author    = {Pumarola, Albert and Corona, Enric and Pons-Moll, Gerard and Moreno-Noguer, Francesc},
  title     = {{D-NeRF}: Neural Radiance Fields for Dynamic Scenes},
  booktitle = {IEEE/CVF Conference on Computer Vision and Pattern Recognition},
  pages     = {10318--10327},
  year      = {2021}
}

@article{park2021hypernerf,
  author  = {Park, Keunhong and Sinha, Utkarsh and Hedman, Peter and Barron, Jonathan T. and Bouaziz, Sofien and Goldman, Dan B. and Martin-Brualla, Ricardo and Seitz, Steven M.},
  title   = {{HyperNeRF}: A Higher-Dimensional Representation for Topologically Varying Neural Radiance Fields},
  journal = {ACM Transactions on Graphics},
  volume  = {40},
  number  = {6},
  pages   = {238:1--238:12},
  year    = {2021},
  doi     = {10.1145/3478513.3480487}
}

@inproceedings{li2023nvfi,
  author    = {Li, Jinxi and Song, Ziyang and Yang, Bo},
  title     = {{NVFi}: Neural Velocity Fields for 3D Physics Learning from Dynamic Videos},
  booktitle = {Advances in Neural Information Processing Systems},
  volume    = {36},
  year      = {2023}
}

@inproceedings{luiten2024dynamic3dgs,
  author    = {Luiten, Jonathon and Kopanas, Georgios and Leibe, Bastian and Ramanan, Deva},
  title     = {Dynamic 3D Gaussians: Tracking by Persistent Dynamic View Synthesis},
  booktitle = {International Conference on 3D Vision},
  pages     = {800--809},
  year      = {2024},
  doi       = {10.1109/3DV62453.2024.00044}
}

@inproceedings{wu2024fourdimensional,
  author    = {Wu, Guanjun and Yi, Taoran and Fang, Jiemin and Xie, Lingxi and Zhang, Xiaopeng and Wei, Wei and Liu, Wenyu and Tian, Qi and Wang, Xinggang},
  title     = {4D Gaussian Splatting for Real-Time Dynamic Scene Rendering},
  booktitle = {IEEE/CVF Conference on Computer Vision and Pattern Recognition},
  pages     = {20310--20320},
  year      = {2024}
}

@inproceedings{yang2024deformable,
  author    = {Yang, Ziyi and Gao, Xinyu and Zhou, Wen and Jiao, Shaohui and Zhang, Yuqing and Jin, Xiaogang},
  title     = {Deformable 3D Gaussians for High-Fidelity Monocular Dynamic Scene Reconstruction},
  booktitle = {IEEE/CVF Conference on Computer Vision and Pattern Recognition},
  pages     = {20331--20341},
  year      = {2024}
}

@inproceedings{li2024spacetime,
  author    = {Li, Zhan and Chen, Zhang and Li, Zhong and Xu, Yi},
  title     = {Spacetime Gaussian Feature Splatting for Real-Time Dynamic View Synthesis},
  booktitle = {IEEE/CVF Conference on Computer Vision and Pattern Recognition},
  pages     = {8508--8520},
  year      = {2024},
  doi       = {10.1109/CVPR52733.2024.00813}
}

@inproceedings{kratimenos2024dynmf,
  author    = {Kratimenos, Agelos and Lei, Jiahui and Daniilidis, Kostas},
  title     = {{DynMF}: Neural Motion Factorization for Real-Time Dynamic View Synthesis with 3D Gaussian Splatting},
  booktitle = {European Conference on Computer Vision},
  pages     = {252--269},
  year      = {2024},
  doi       = {10.1007/978-3-031-72986-7_15}
}

@inproceedings{lei2025mosca,
  author    = {Lei, Jiahui and Weng, Yijia and Harley, Adam W. and Guibas, Leonidas and Daniilidis, Kostas},
  title     = {{MoSca}: Dynamic Gaussian Fusion from Casual Videos via 4D Motion Scaffolds},
  booktitle = {IEEE/CVF Conference on Computer Vision and Pattern Recognition},
  pages     = {6165--6177},
  year      = {2025}
}

@inproceedings{zhao2024gaussianprediction,
  author    = {Zhao, Boming and Li, Yuan and Sun, Ziyu and Zeng, Lin and Shen, Yujun and Ma, Rui and Zhang, Yinda and Bao, Hujun and Cui, Zhaopeng},
  title     = {{GaussianPrediction}: Dynamic 3D Gaussian Prediction for Motion Extrapolation and Free View Synthesis},
  booktitle = {ACM SIGGRAPH Conference Papers},
  pages     = {84:1--84:12},
  year      = {2024},
  doi       = {10.1145/3641519.3657417}
}

@inproceedings{wang2026odegs,
  author    = {Wang, Daniel and Rim, Patrick and Tian, Tian and Lao, Dong and Wong, Alex and Sundaramoorthi, Ganesh},
  title     = {{ODE-GS}: Latent {ODE}s for Dynamic Scene Extrapolation with 3D Gaussian Splatting},
  booktitle = {International Conference on Learning Representations},
  year      = {2026}
}

@InProceedings{Lee_2026_CVPR,
    author    = {Lee, Junmyeong and Choi, Hoseung and Cho, Minsu},
    title     = {Space-Time Forecasting of Dynamic Scenes with Motion-aware Gaussian Grouping},
    booktitle = {Proceedings of the IEEE/CVF Conference on Computer Vision and Pattern Recognition (CVPR)},
    month     = {June},
    year      = {2026},
    pages     = {41033-41043}
}

@inproceedings{jin2026moegs,
  author    = {Jin, In-Hwan and Mun, Hyeongju and Kim, Joonsoo and Yun, Kugjin and Kong, Kyeongbo},
  title     = {{MoE-GS}: Mixture of Experts for Dynamic Gaussian Splatting},
  booktitle = {International Conference on Learning Representations},
  year      = {2026},
  note      = {arXiv:2510.19210}
}

@inproceedings{zhu2024fsgs,
  title     = {{FSGS}: Real-Time Few-Shot View Synthesis Using Gaussian
               Splatting},
  author    = {Zhu, Zehao and Fan, Zhiwen and Jiang, Yifan and Wang, Zhangyang},
  booktitle = {Proceedings of the European Conference on Computer Vision},
  year      = {2024},
  doi       = {10.1007/978-3-031-72933-1_9}
}

@article{salehi2026lgdwtgs,
  author  = {Shima Salehi and Atharva Agashe and
             Andrew J. McFarland and Joshua Peeples},
  title   = {{LGDWT-GS}: Local and Global Discrete
             Wavelet-Regularized 3-D Gaussian Splatting
             for Sparse-View Scene Reconstruction},
  journal = {IEEE Access},
  volume  = {14},
  pages   = {72746--72756},
  year    = {2026},
  doi     = {10.1109/ACCESS.2026.3691681}
}

@inproceedings{vaswani2017attention,
  title     = {Attention Is All You Need},
  author    = {Vaswani, Ashish and Shazeer, Noam and Parmar, Niki
               and Uszkoreit, Jakob and Jones, Llion and Gomez,
               Aidan N. and Kaiser, Lukasz and Polosukhin, Illia},
  booktitle = {Advances in Neural Information Processing Systems},
  year      = {2017}
}

@article{ahmed1974dct,
  title   = {Discrete Cosine Transform},
  author  = {Ahmed, Nasir and Natarajan, T. and Rao, K. R.},
  journal = {IEEE Transactions on Computers},
  volume  = {C-23},
  number  = {1},
  pages   = {90--93},
  year    = {1974}
}

@article{kabsch1976solution,
  title   = {A Solution for the Best Rotation to Relate Two Sets of Vectors},
  author  = {Kabsch, Wolfgang},
  journal = {Acta Crystallographica Section A},
  volume  = {32},
  number  = {5},
  pages   = {922--923},
  year    = {1976}
}

@inproceedings{hu2021fiery,
  author    = {Hu, Anthony and Murez, Zak and Mohan, Nikhil and
               Dudas, Sof{\'i}a and Hawke, Jeffrey and
               Badrinarayanan, Vijay and Cipolla, Roberto and
               Kendall, Alex},
  title     = {{FIERY}: Future Instance Prediction in Bird's-Eye View
               from Surround Monocular Cameras},
  booktitle = {Proceedings of the IEEE/CVF International Conference
               on Computer Vision},
  pages     = {15253--15262},
  year      = {2021},
  doi       = {10.1109/ICCV48922.2021.01499}
}

@inproceedings{yang2023unisim,
  author    = {Yang, Ze and Chen, Yun and Wang, Jingkang and
               Manivasagam, Sivabalan and Ma, Wei-Chiu and
               Yang, Anqi Joyce and Urtasun, Raquel},
  title     = {{UniSim}: A Neural Closed-Loop Sensor Simulator},
  booktitle = {Proceedings of the IEEE/CVF Conference on Computer
               Vision and Pattern Recognition},
  pages     = {1389--1399},
  year      = {2023},
  doi       = {10.1109/CVPR52729.2023.00140}
}

@inproceedings{zheng2024occworld,
  author    = {Zheng, Wenzhao and Chen, Weiliang and Huang, Yuanhui and
               Zhang, Borui and Duan, Yueqi and Lu, Jiwen},
  title     = {{OccWorld}: Learning a {3D} Occupancy World Model for
               Autonomous Driving},
  booktitle = {European Conference on Computer Vision},
  pages     = {55--72},
  year      = {2024},
  doi       = {10.1007/978-3-031-72624-8_4}
}

@inproceedings{kayalibay2022spatial,
  author    = {Kayalibay, Baris and Mirchev, Atanas and
               van der Smagt, Patrick and Bayer, Justin},
  title     = {Tracking and Planning with Spatial World Models},
  booktitle = {Proceedings of the 4th Annual Learning for Dynamics
               and Control Conference},
  series    = {Proceedings of Machine Learning Research},
  volume    = {168},
  pages     = {124--137},
  publisher = {PMLR},
  year      = {2022},
  url       = {https://proceedings.mlr.press/v168/kayalibay22a.html}
}

@inproceedings{ost2021neural,
  author    = {Ost, Julian and Mannan, Fahim and Thuerey, Nils and
               Knodt, Julian and Heide, Felix},
  title     = {Neural Scene Graphs for Dynamic Scenes},
  booktitle = {Proceedings of the IEEE/CVF Conference on Computer
               Vision and Pattern Recognition},
  pages     = {2856--2865},
  year      = {2021}
}
}

\clearpage
\thispagestyle{empty}
\null
\clearpage
\appendix
\twocolumn[
\begin{center}
    {\LARGE\bfseries
    MoRE-GS: Motion-Regime Experts for Future Gaussian Splatting\par}
    \vspace{0.5em}
    {\Large\bfseries Supplementary Material\par}
    \vspace{0.8em}
    {\large Shima Salehi \qquad Joshua Peeples\par}
    \vspace{0.3em}
    Department of Electrical and Computer Engineering\\
    Texas A\&M University, College Station, Texas, USA\\
    {\small\texttt{shima.salehi@tamu.edu \qquad jpeeples@tamu.edu}}
\end{center}
\vspace{1em}
]


\captionsetup[table]{font=small,skip=3pt}
\newcommand{\suppTableStyle}{%
    \scriptsize
    \setlength{\tabcolsep}{1.8pt}%
    \renewcommand{\arraystretch}{1.08}%
}
\newcommand{\suppWideTableStyle}{%
    \small
    \setlength{\tabcolsep}{4.0pt}%
    \renewcommand{\arraystretch}{1.08}%
}

\section{Protocol and Scope}
\label{sec:supp-protocol}

All experiments use chronological observed/future splits. The dynamic Gaussian
representation, control keypoints, keypoint-to-Gaussian bindings, appearance,
opacity, and renderer are fitted only on the observed prefix and frozen before
forecasting. Router calibration further divides the observed prefix 75/25:
temporary experts fit the first part and predict the disjoint observed tail;
the calibrated router is then frozen and paired with deployment experts trained
on the complete observed prefix. No future frame or metric is used for
reconstruction, grouping, expert fitting, routing, checkpoint selection, or
early stopping. The calibration tail is excluded only from temporary
motion-expert fitting; it remains part of the frozen reconstruction. Thus,
calibration is motion-expert-out-of-fit. A
fit-only reconstruction would change the underlying representation and is not
treated as a matched router ablation. Every reported checkpoint is cold reloaded before evaluation.

The primary D-NeRF~\cite{pumarola2021dnerf} protocol uses an 80/20 split.
Forecasting stability is evaluated with expert/router seeds 17, 42, and 123;
all other supplementary analyses use seed 17. Split-sensitivity experiments
use independently reconstructed 70/30 and 90/10 representations. All runs use
$G=16$ soft motion groups, four horizon bins, and a 256-parameter route table.
Peak signal-to-noise ratio (PSNR), structural similarity index measure (SSIM),
and learned perceptual image patch similarity (LPIPS) are averaged within a
scene and then equally across scenes. Exact implementation settings are given
below.

\section{Additional Implementation Details}
\label{sec:supp-implementation}

This section reports only settings not specified in the main paper. Soft-group
assignment uses temperature $0.18$ and 30 deterministic $k$-means updates. The
coefficient Transformer uses AdamW with weight decay $10^{-4}$; the structured
Transformer uses Adam. Both use cosine learning-rate decay. The router uses
zero weight decay and gradient-norm clipping at 1.0.

For the important-pixel mask defined in the main paper, the foreground and
dynamic-change thresholds are $\tau_{\rm fg}=0.04$ and
$\tau_{\rm dyn}=0.025$, respectively. The masked mean-squared-error term has
weight 1.0. The adjacent-bin smoothness, moving-mass, and moving-expert entropy
weights are $\lambda_s=2\times10^{-3}$, $\lambda_m=10^{-4}$, and
$\lambda_e=10^{-4}$, respectively.

\section{Router Visualization}
\label{sec:supp-router-vis}

\begin{figure*}[t]
\centering
\includegraphics[width=0.97\textwidth]{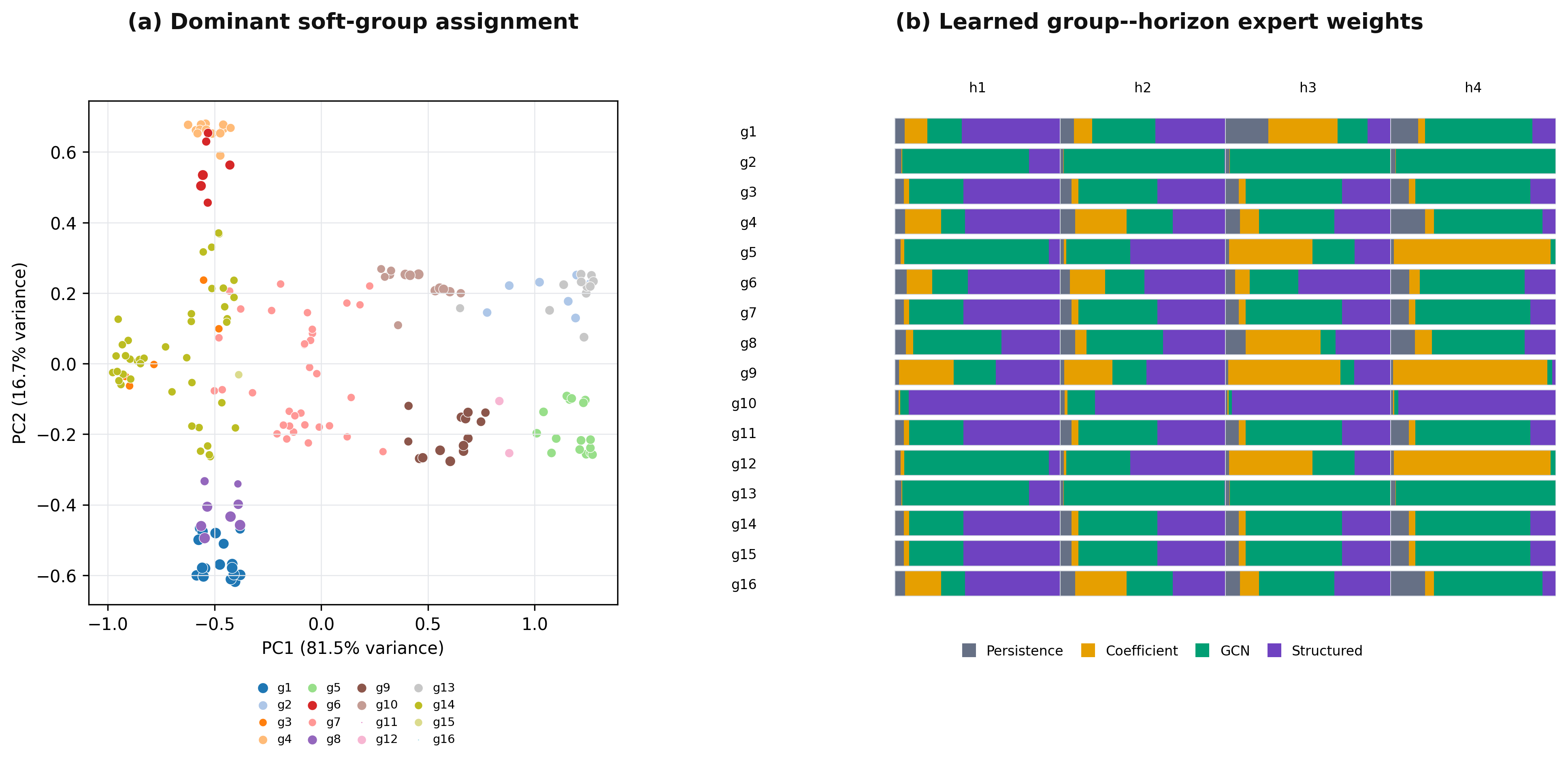}
\caption{Router visualization for $G=16$: (a) dominant soft-group assignment,
with color indicating group and marker size indicating confidence; inference
retains full soft assignments. (b) Group--horizon expert weights.}
\label{fig:supp-routes}
\end{figure*}

Figure~\ref{fig:supp-routes} visualizes the saved Jumping Jacks router using
observed-prefix information only. For visualization, the left panel colors each
keypoint by its maximum-probability group and scales its marker by assignment
confidence in a two-dimensional principal-component projection. This hard
color label is used only for visualization; inference retains the complete
soft assignment vector. The right panel shows the learned expert weights for
every group and horizon bin.

\begin{figure*}[!t]
\centering
\includegraphics[width=0.98\textwidth]{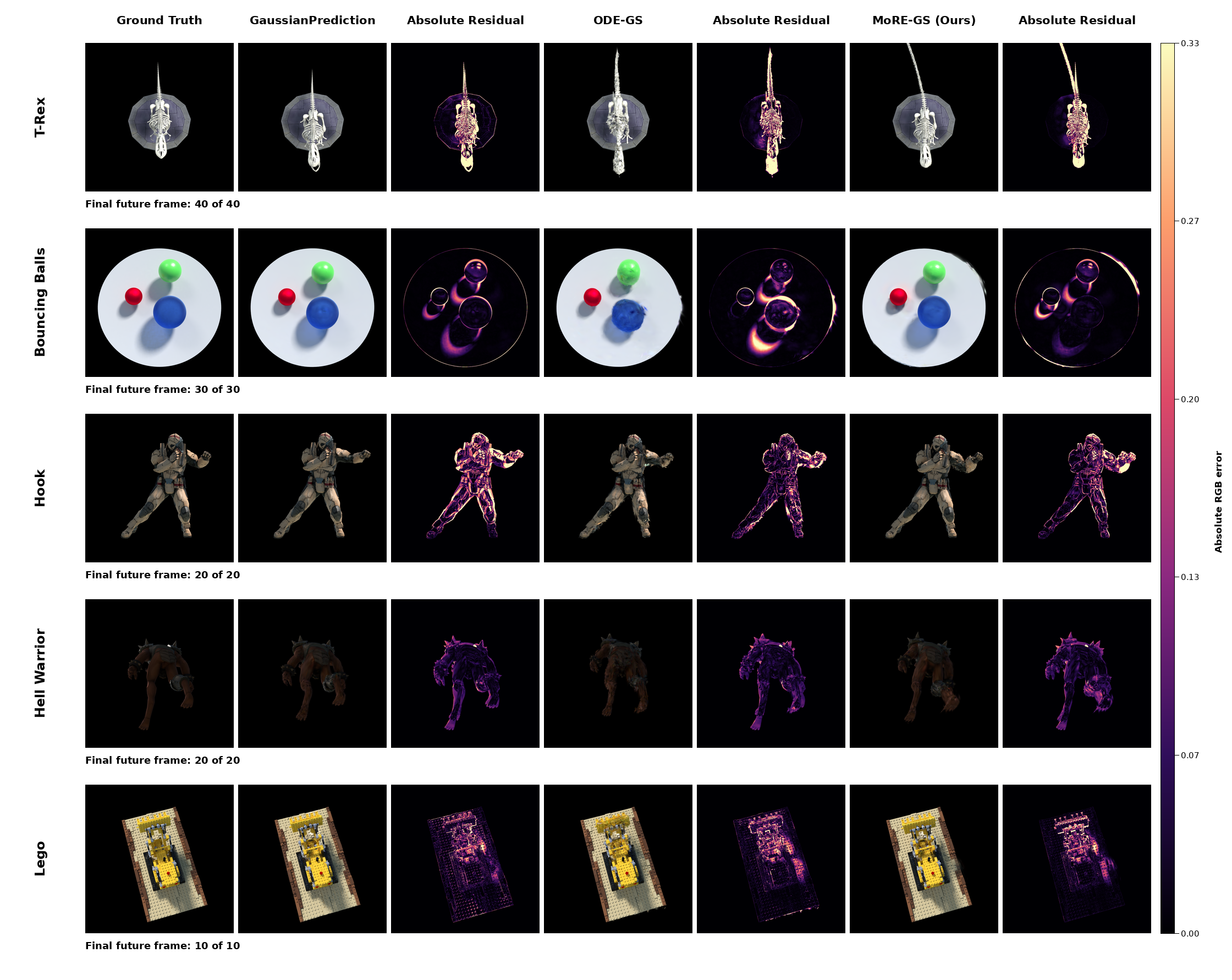}
\caption{Final-horizon forecasts and absolute RGB residuals on the five
D-NeRF scenes not visualized in the main paper.}
\label{fig:supp-additional-scenes}
\end{figure*}
\section{Forecasting Stability}
\label{sec:supp-stability}

Table~\ref{tab:supp-stability} evaluates three expert/router initialization
seeds while holding the frozen Gaussian reconstruction fixed. Thus, the
variation measures forecasting-head optimization stability rather than
reconstruction variability. The scene-unweighted mean is
$27.110\!\pm\!0.279$ dB. Relative to persistence, the three-seed model gains
2.409 dB; a scene-resampled bootstrap gives a 95\% confidence interval of
$[0.883,4.173]$ dB, and an exact one-sided paired sign-flip test gives
$p=0.0039$.

\begin{table}[!t]
\centering
\caption{$G=16$ stability over seeds 17, 42, and 123; values are
mean $\pm$ standard deviation.}
\label{tab:supp-stability}
\suppTableStyle
\begin{tabular*}{\columnwidth}{
@{\extracolsep{\fill}}lccc@{}}
\toprule
Scene & PSNR $\uparrow$ & SSIM $\uparrow$ & LPIPS $\downarrow$ \\
\midrule
T-Rex
& $20.61\!\pm\!0.64$
& $0.934\!\pm\!0.012$
& $0.059\!\pm\!0.012$ \\

Mutant
& $33.19\!\pm\!0.27$
& $0.975\!\pm\!0.001$
& $0.015\!\pm\!<0.001$ \\

Jumping Jacks
& $21.20\!\pm\!0.10$
& $0.922\!\pm\!0.003$
& $0.072\!\pm\!0.002$ \\

Stand Up
& $30.36\!\pm\!0.23$
& $0.956\!\pm\!0.001$
& $0.035\!\pm\!0.001$ \\

Bouncing Balls
& $27.60\!\pm\!0.57$
& $0.972\!\pm\!0.003$
& $0.037\!\pm\!0.005$ \\

Hook
& $26.50\!\pm\!0.07$
& $0.938\!\pm\!<0.001$
& $0.041\!\pm\!<0.001$ \\

Hell Warrior
& $30.54\!\pm\!0.27$
& $0.922\!\pm\!0.003$
& $0.078\!\pm\!0.003$ \\

Lego
& $26.89\!\pm\!0.95$
& $0.945\!\pm\!0.003$
& $0.043\!\pm\!0.003$ \\
\midrule
\textbf{Mean}
& $\mathbf{27.11\!\pm\!0.28}$
& $\mathbf{0.946\!\pm\!0.003}$
& $\mathbf{0.047\!\pm\!0.003}$ \\
\bottomrule
\end{tabular*}
\end{table}

\section{Horizon and Quartile Analysis}
\label{sec:supp-horizon}

Each sealed 80/20 suffix is divided chronologically into four approximately
equal quartiles. Table~\ref{tab:supp-persistence-quartiles} compares MoRE-GS
with persistence. PSNR improves by 3.12, 3.17, 3.02, and 1.73 dB from Q1 to
Q4, respectively, with better SSIM and LPIPS throughout. The smaller Q4 margin
indicates increasing long-horizon uncertainty.

Table~\ref{tab:supp-routing-quartiles} compares the group-only run corresponding
to the 27.38 dB ablation result with the selected group--horizon model. Although
group--horizon routing is 0.45 dB lower in Q1, it improves PSNR by 0.10, 0.47,
and 0.33 dB in Q2--Q4, respectively. It also improves full-suffix PSNR by
0.089 dB, SSIM by 0.0020, and LPIPS by 0.0015. We therefore retain
group--horizon routing because it adapts expert selection as extrapolation
uncertainty increases and provides more reliable performance at the longer
forecast horizons.

\begin{table*}[!t]
\centering
\caption{Quartile forecasts for persistence and MoRE-GS. Best values
per row and metric are bold.}
\label{tab:supp-persistence-quartiles}
\suppWideTableStyle
\begin{tabular*}{\textwidth}{
@{\extracolsep{\fill}}l*{6}{c}@{}}
\toprule
& \multicolumn{3}{c}{Persistence}
& \multicolumn{3}{c}{MoRE-GS} \\
\cmidrule(lr){2-4}
\cmidrule(lr){5-7}
Quartile
& PSNR$\uparrow$ & SSIM$\uparrow$ & LPIPS$\downarrow$
& PSNR$\uparrow$ & SSIM$\uparrow$ & LPIPS$\downarrow$ \\
\midrule
Q1
& 28.31 & 0.955 & 0.032
& \textbf{31.43} & \textbf{0.970} & \textbf{0.024} \\

Q2
& 24.51 & 0.937 & 0.050
& \textbf{27.68} & \textbf{0.951} & \textbf{0.040} \\

Q3
& 23.23 & 0.929 & 0.061
& \textbf{26.25} & \textbf{0.941} & \textbf{0.050} \\

Q4
& 22.39 & 0.921 & 0.070
& \textbf{24.12} & \textbf{0.928} & \textbf{0.065} \\
\bottomrule
\end{tabular*}
\end{table*}

\begin{table*}[!t]
\centering
\caption{Quartile forecasts for group-only and group--horizon routing.
Best values per row and metric are bold.}
\label{tab:supp-routing-quartiles}
\suppWideTableStyle
\begin{tabular*}{\textwidth}{
@{\extracolsep{\fill}}l*{6}{c}@{}}
\toprule
& \multicolumn{3}{c}{Group-only}
& \multicolumn{3}{c}{Group--horizon} \\
\cmidrule(lr){2-4}
\cmidrule(lr){5-7}
Quartile
& PSNR$\uparrow$ & SSIM$\uparrow$ & LPIPS$\downarrow$
& PSNR$\uparrow$ & SSIM$\uparrow$ & LPIPS$\downarrow$ \\
\midrule
Q1
& \textbf{31.88} & 0.968 & \textbf{0.024}
& 31.43 & \textbf{0.970} & \textbf{0.024} \\

Q2
& 27.58 & 0.950 & 0.042
& \textbf{27.68} & \textbf{0.951} & \textbf{0.040} \\

Q3
& 25.77 & 0.938 & 0.053
& \textbf{26.25} & \textbf{0.941} & \textbf{0.050} \\

Q4
& 23.79 & 0.926 & 0.067
& \textbf{24.12} & \textbf{0.928} & \textbf{0.065} \\
\midrule
Mean
& 27.38 & 0.946 & 0.046
& \textbf{27.47} & \textbf{0.948} & \textbf{0.045} \\
\bottomrule
\end{tabular*}
\end{table*}

\section{Temporal-Basis Ablation}
\label{sec:supp-temporal-basis}

Table~\ref{tab:supp-basis} compares the random orthonormal basis with
DCT-II~\cite{ahmed1974dct} while retaining all coefficients and matching all
other settings. The random basis improves mean PSNR by 0.185 dB. We retain it
because it preserves trajectory information without imposing DCT-II's fixed
frequency ordering and achieves better aggregate performance across all three
metrics.

\begin{table*}[!t]
\centering
\caption{Matched temporal-basis ablation. Best values per scene and
metric are bold.}
\label{tab:supp-basis}
\suppWideTableStyle
\begin{tabular*}{\textwidth}{
@{\extracolsep{\fill}}l*{6}{c}@{}}
\toprule
& \multicolumn{3}{c}{DCT-II}
& \multicolumn{3}{c}{Random orthonormal} \\
\cmidrule(lr){2-4}
\cmidrule(lr){5-7}
Scene
& PSNR$\uparrow$ & SSIM$\uparrow$ & LPIPS$\downarrow$
& PSNR$\uparrow$ & SSIM$\uparrow$ & LPIPS$\downarrow$ \\
\midrule
T-Rex
& 19.04 & 0.924 & 0.067
& \textbf{19.05} & \textbf{0.930} & \textbf{0.063} \\

Mutant
& \textbf{32.70} & \textbf{0.969} & \textbf{0.016}
& 32.31 & 0.968 & 0.017 \\

Jumping Jacks
& 20.61 & 0.907 & 0.086
& \textbf{20.82} & \textbf{0.910} & \textbf{0.084} \\

Stand Up
& 28.92 & 0.951 & 0.040
& \textbf{29.37} & \textbf{0.953} & \textbf{0.038} \\

Bouncing Balls
& 26.47 & 0.970 & 0.043
& \textbf{26.99} & \textbf{0.973} & \textbf{0.039} \\

Hook
& 26.45 & 0.940 & 0.040
& \textbf{26.97} & \textbf{0.941} & \textbf{0.039} \\

Hell Warrior
& 29.60 & 0.902 & 0.095
& \textbf{30.17} & \textbf{0.918} & \textbf{0.082} \\

Lego
& \textbf{25.30} & \textbf{0.938} & \textbf{0.050}
& 24.90 & 0.936 & 0.052 \\
\midrule
Mean
& 26.14 & 0.938 & 0.055
& \textbf{26.32} & \textbf{0.941} & \textbf{0.052} \\
\bottomrule
\end{tabular*}
\end{table*}


\section{Observed/Future Split Robustness}
\label{sec:supp-splits}

We compare MoRE-GS with
GaussianPrediction~\cite{zhao2024gaussianprediction} under matched 70/30 and
90/10 conditions. Each split uses its own observed-only Gaussian
reconstruction, identical future indices for both forecasters, forecasting
seed 17, and all eight scenes. Tables~\ref{tab:supp-split-7030} and
\ref{tab:supp-split-9010} report all scene-level metrics.

MoRE-GS improves every aggregate metric, with PSNR gains of 2.066 dB at 70/30
and 1.635 dB at 90/10. GaussianPrediction achieves higher PSNR on Jumping
Jacks under both splits and on Hell Warrior at 90/10. MoRE-GS achieves higher
PSNR on seven of eight scenes at 70/30 and five of eight scenes at 90/10,
with an additional tie on T-Rex at 90/10. Changing the split affects both the
available reconstruction evidence and the forecast length.

\begin{table*}[!t]
\centering
\caption{D-NeRF forecasting under the matched 70/30 observed/future split.
Best values per scene and metric are bold.}
\label{tab:supp-split-7030}
\suppWideTableStyle
\begin{tabular*}{\textwidth}{
@{\extracolsep{\fill}}l*{6}{c}@{}}
\toprule
& \multicolumn{3}{c}{GaussianPrediction}
& \multicolumn{3}{c}{MoRE-GS} \\
\cmidrule(lr){2-4}
\cmidrule(lr){5-7}
Scene
& PSNR$\uparrow$ & SSIM$\uparrow$ & LPIPS$\downarrow$
& PSNR$\uparrow$ & SSIM$\uparrow$ & LPIPS$\downarrow$ \\
\midrule
T-Rex
& 19.52 & \textbf{0.933} & \textbf{0.063}
& \textbf{19.64} & 0.929 & 0.067 \\

Mutant
& 25.31 & 0.941 & 0.037
& \textbf{28.97} & \textbf{0.957} & \textbf{0.024} \\

Jumping Jacks
& \textbf{19.32} & \textbf{0.907} & \textbf{0.095}
& 18.53 & 0.837 & 0.153 \\

Stand Up
& 22.92 & 0.926 & 0.068
& \textbf{27.60} & \textbf{0.945} & \textbf{0.044} \\

Bouncing Balls
& 23.17 & 0.955 & 0.052
& \textbf{27.73} & \textbf{0.977} & \textbf{0.032} \\

Hook
& 22.74 & 0.902 & 0.068
& \textbf{24.52} & \textbf{0.919} & \textbf{0.059} \\

Hell Warrior
& 28.52 & \textbf{0.904} & \textbf{0.098}
& \textbf{28.82} & 0.900 & 0.101 \\

Lego
& 23.12 & 0.915 & 0.069
& \textbf{25.33} & \textbf{0.932} & \textbf{0.057} \\
\midrule
Mean
& 23.08 & 0.923 & 0.069
& \textbf{25.14} & \textbf{0.924} & \textbf{0.067} \\
\bottomrule
\end{tabular*}
\end{table*}

\begin{table*}[!t]
\centering
\caption{D-NeRF forecasting under the matched 90/10 observed/future split.
Best values per scene and metric are bold.}
\label{tab:supp-split-9010}
\suppWideTableStyle
\begin{tabular*}{\textwidth}{
@{\extracolsep{\fill}}l*{6}{c}@{}}
\toprule
& \multicolumn{3}{c}{GaussianPrediction}
& \multicolumn{3}{c}{MoRE-GS} \\
\cmidrule(lr){2-4}
\cmidrule(lr){5-7}
Scene
& PSNR$\uparrow$ & SSIM$\uparrow$ & LPIPS$\downarrow$
& PSNR$\uparrow$ & SSIM$\uparrow$ & LPIPS$\downarrow$ \\
\midrule
T-Rex
& \textbf{22.62} & 0.952 & \textbf{0.039}
& \textbf{22.62} & \textbf{0.954} & \textbf{0.039} \\

Mutant
& 27.90 & 0.956 & 0.023
& \textbf{30.39} & \textbf{0.966} & \textbf{0.017} \\

Jumping Jacks
& \textbf{23.00} & \textbf{0.938} & \textbf{0.051}
& 21.87 & 0.928 & 0.064 \\

Stand Up
& 30.75 & 0.964 & 0.024
& \textbf{32.22} & \textbf{0.971} & \textbf{0.020} \\

Bouncing Balls
& 29.90 & 0.980 & 0.020
& \textbf{34.04} & \textbf{0.988} & \textbf{0.015} \\

Hook
& 25.86 & 0.933 & 0.037
& \textbf{28.66} & \textbf{0.957} & \textbf{0.026} \\

Hell Warrior
& \textbf{33.59} & \textbf{0.952} & \textbf{0.049}
& 33.13 & 0.946 & 0.056 \\

Lego
& 26.39 & 0.929 & 0.044
& \textbf{30.15} & \textbf{0.961} & \textbf{0.031} \\
\midrule
Mean
& 27.50 & 0.951 & 0.036
& \textbf{29.13} & \textbf{0.959} & \textbf{0.033} \\
\bottomrule
\end{tabular*}
\end{table*}

\section{Additional Qualitative Results}
\label{sec:supp-additional-qualitative}

Figure~\ref{fig:supp-additional-scenes} complements the main-paper qualitative
comparison with the five D-NeRF scenes not visualized there. Mutant, Stand Up,
and Jumping Jacks are omitted because they already appear in the main paper.
All residuals use one shared display scale. Future ground truth is used only
for this post-evaluation visualization and residual computation.

\end{document}